\documentclass[journal]{IEEEtran}
\usepackage[keeplastbox]{flushend}
\usepackage{color} 
\usepackage{epsfig}
\usepackage{graphicx}
\usepackage{amsmath}
\usepackage{amssymb}
\usepackage{amsfonts}
\usepackage{amsthm}
\usepackage{dsfont}
\usepackage{bm}
\usepackage{bbm}
\usepackage{url}
\usepackage{cases}
\usepackage{subeqnarray}
\usepackage{mathrsfs}
\usepackage{booktabs}
\usepackage{threeparttable}
\usepackage{enumerate}
\usepackage{flushend}
\usepackage{ragged2e}
\usepackage{mathrsfs}
\usepackage{algorithm}
\usepackage{algorithmic}
\usepackage{todonotes}
\usepackage{fancyhdr}
\usepackage{subfigure}
\usepackage{balance}

\def\ie{{\em i.e.}}
\def\eg{{\em e.g.}}
\def\etal{{\em et al.}}
\def\wrt{{\em w.r.t.}}

\begin{document}
\title{Image Co-segmentation via Multi-scale \\Local Shape Transfer}

\author{Wei~Teng,~
	Yu~Zhang,~
	Xiaowu~Chen,~
	Jia~Li,~
	and~Zhiqiang~He
	
	\thanks{W. Teng, Y. Zhang, X. Chen and J. Li are with the State Key Laboratory of Virtual Reality Technology and Systems, School of Computer Science and Engineering, Beihang University, Beijing 100191, China. (e-mail: tengw@buaa.edu.cn; zhangyulb@gmail.com; chen@buaa.edu.cn; jiali@buaa.edu.cn)}
	
	\thanks{J. Li is also with the International Research Institute for Multidisciplinary Science, Beihang University, Beijing 100191, China.}
	
	\thanks{Z. He is with the Lenovo Research, China. (e-mail: lirong2@lenovo.com)}
	\thanks{\emph{Corresponding author: Xiaowu Chen (e-mail: chen@buaa.edu.cn).} }
	\thanks{\emph{Earlier version of this work has been published in BMVC 2016~\cite{dlle}.}}
}

\maketitle

\begin{abstract}
Image co-segmentation is a challenging task in computer vision that aims to segment all pixels of the objects from a predefined semantic category. In real-world cases, however, common foreground objects often vary greatly in appearance, making their global shapes
highly inconsistent across images and difficult to be segmented. To address this problem, this paper proposes a novel co-segmentation approach that transfers patch-level local object shapes which appear more consistent across different images. In our framework,
a multi-scale patch neighbourhood system is first generated using proposal flow on arbitrary image-pair, which is further refined by Locally Linear Embedding. Based on the patch relationships, we propose an efficient algorithm to jointly segment the objects in each image while transferring their local shapes across different images. Extensive experiments demonstrate that the proposed method can robustly and effectively segment common objects from an image set. 
On iCoseg, MSRC and Coseg-Rep dataset, the proposed approach performs comparable or better than the state-of-the-arts, while on a more challenging benchmark Fashionista dataset, our method achieves significant improvements.
\end{abstract}

\begin{IEEEkeywords}
Image co-segmentation, Shape transfer, Locally linear embedding.
\end{IEEEkeywords}

\IEEEpeerreviewmaketitle

\section{Introduction}

\IEEEPARstart{i}{mage} co-segmentation~(\eg~\cite{discriminative,short_path-seg,interactive-seg,co-fusion,QUANCVPR2016}) is an important problem in the field of computer vision and multimedia.
It aims to simultaneously segment all the pixels of the common foreground object(s) from the same category in a set of images.
With its rapid development, image co-segmentation has supported various vision applications, such as 
fine-grained object recognition~\cite{fine_grained}, visual concept discovery~\cite{visual_knowledge},
and image retrieval~\cite{tmm_imageretrieval}, 
which greatly facilitated automatic analysis and utilization of large-scale multimedia data.

\begin{figure*}
	\centering
	\includegraphics[width=1\textwidth]{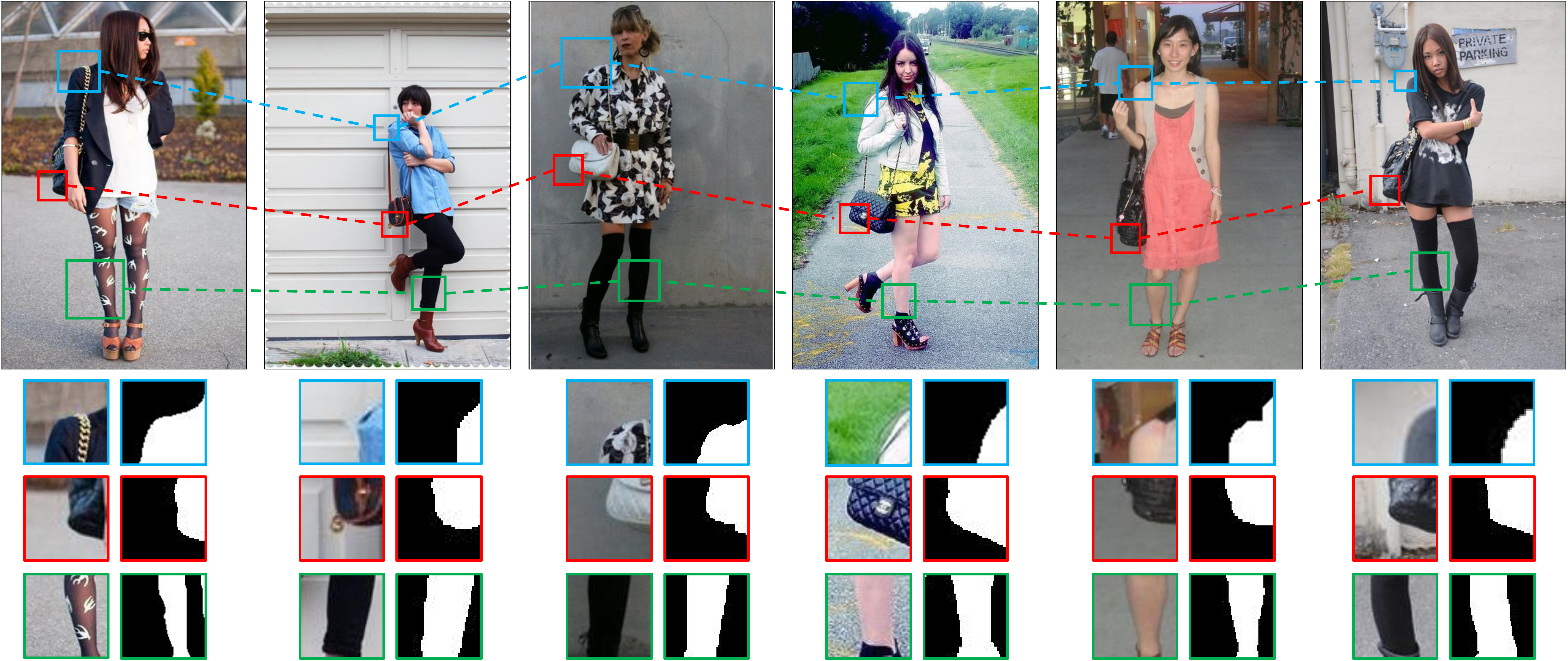}
	\caption{Illustration of the local shape consistency. Holistic shape of the object in examplar images varies due to pose differences. However, object shape in local patches~(displayed in bounding boxes with the same color) remain similar across different images.}
	\label{fig:intro}
\end{figure*}  
In the past decade, various visual cues have been investigated for extracting objects from a predefined semantic image set.
Among them, appearance cues such as color and texture have been proved effective. 
Based on the appearance descriptor, the common objects can be discovered by either learning a shared distribution~\cite{discriminative} or constructing correspondences among them~\cite{fmaps}.
However, appearance cues are limited in distinguishing object from visually similar background, and dealing with complex object.

To address these problems, some studies have turned to the shape cues for reducing foreground or background ambiguities~\cite{patch_cut}. 
One popular approach is learning a shared shape model across different images. Exemplar works include shape priors~\cite{classcut}, ~\cite{segmentation_imagenet}, and deformable shape templates~\cite{cosketch}. However, these approaches often have difficulty segmenting the common objects in such scenarios with large variance of viewpoints, scales and object poses. Moreover, they often require learning a uniform objects model, which may be unobtainable when the global shapes of those common objects are inconsistent. 
As a result, these template-based approaches
may not work well.

By observing the common objects at patch level, we find that no matter how holistic shape varies, local shapes are stable and thus transferable~(see Fig.~\ref{fig:intro}).
Inspired by this observation, we propose a novel framework for image co-segmentation by introducing patch-level shape transfer.
Specifically, we first generate patches via multi-scale sliding windows.
For each patch, we search for its transferable neighbours in other homogeneous images. 
To prune out the unreliable
matches, we further learn a sparser neighbourhood system for the image patch set using Locally
Linear Embedding~\cite{lle}. 
Then a novel image co-segmentation approach is proposed by introducing patch-level consistencies into a graph-cut based segmentation framework.
Extensive experiments on iCoseg~\cite{icoseg}, MSRC~\cite{MSRC} and Coseg-rep~\cite{cosketch} dataset show that our approach performs comparable or better than the state-of-the-arts.
Moreover, on the challenging benchmark Fasionista dataset~\cite{parsing_clothing} with complex object appearances and poses, the proposed approach achieves remarkable improvements.

Our main contributions are summarized as follows:

1) we propose a novel image co-segmentation framework by introducing multi-scale local
shape transfer.

2) we present a strategy to refine the patch correspondences
in an image set through Locally Linear Embedding.

3) we propose an efficient co-segmentation algorithm by embedding patch
consistencies into graph-cut based energy.

This work extends our previous study~\cite{dlle} in two aspects:
1) Inspired by the success of multi-scale strategy in many vision and multimedia tasks (\eg~\cite{yonghong_2013, tmm_multiscale,MAPCVPR2016,tian_ijcv, meng_tip16}), 
we associate local patches across images at different resolutions via multi-scale sliding window.

2) we conduct more comparisons between our approach and state-of-the-arts methods on iCoseg dataset~\cite{icoseg}, MSRC dataset~\cite{MSRC}, Coseg-Rep dataset~\cite{cosketch} and Fashionista dataset~\cite{parsing_clothing}. Quantitative results show that
the proposed algorithm achieves significant improvements
in both accuracy and speed.

In the rest of this paper, Section \uppercase\expandafter{\romannumeral2} briefly reviews
the previous researches on image co-segmentation and Section \uppercase\expandafter{\romannumeral3} describes  the proposed image co-segmentation framework in detail. Experimental results are presented in Section \uppercase\expandafter{\romannumeral4}. Finally, the paper is concluded in Section \uppercase\expandafter{\romannumeral5}.

\section{Related Works}
Existing studies on image co-segmentation algorithms can be roughly  divided into categories: template-based and matching-based groups, which will be introduced in Section \uppercase\expandafter{\romannumeral2-A} and Section \uppercase\expandafter{\romannumeral2-B} respectively.
Relavant works on shape transfer is also presented in Section \uppercase\expandafter{\romannumeral2-C}.

\subsection{Template-based Group}
Most template-based approaches assume that there exists a single model which can be generalized to represent all the objects in a specific image set. Following this idea, some works proposed to learn shared distributions of appearance features. For example, 
Jolin \etal~\cite{discriminative} combined spectral clustering and kernel methods into a discriminative clustering framework, which learned linear models jointly for foreground and background based on color and texture features.
Kim \etal~\cite{KimICCV11} modeled the co-segmentation as a temperature maximization problem of anisotropic heat diffusion, in which foreground objects were represented with a diffusion process. However, these models are often insufficient for objects with complex appearances. 
To address this issue, several works such as~\cite{classcut, segmentation_imagenet, cosketch, sa_coseg} advocated using shape models to facilitate co-segmentation. A common practice is to learn shape prior maps, which indicate the likelihoods of the common objects appearing at different image locations. As object shapes are actually unknown in co-segmentation, the shape priors were iteratively refined using the current segmentations~\cite{segmentation_imagenet,classcut}. In~\cite{cosketch}, a sophisticated model was designed to jointly segment the common objects and learn their deformable shape templates. 
In~\cite{sa_coseg}, a common shape pattern was discovered through Affinity Propagation~(AP) clustering to refine the segmentations of image sets.
More recently, Quan \etal~\cite{QUANCVPR2016} established close-loop graph to represent the foreground and background separately, and they applied both low-level appearance and high-level semantic features.

To sum up, template-based approaches are powerful since they output not only the segmented objects but also the learned foreground/background/shape models. However, the models proposed are often simple in tractability during learning or inference, and thus may not adequately capture real-world object with various appearances and structures.

\begin{figure*}
	\centering
	\includegraphics[width=1\textwidth]{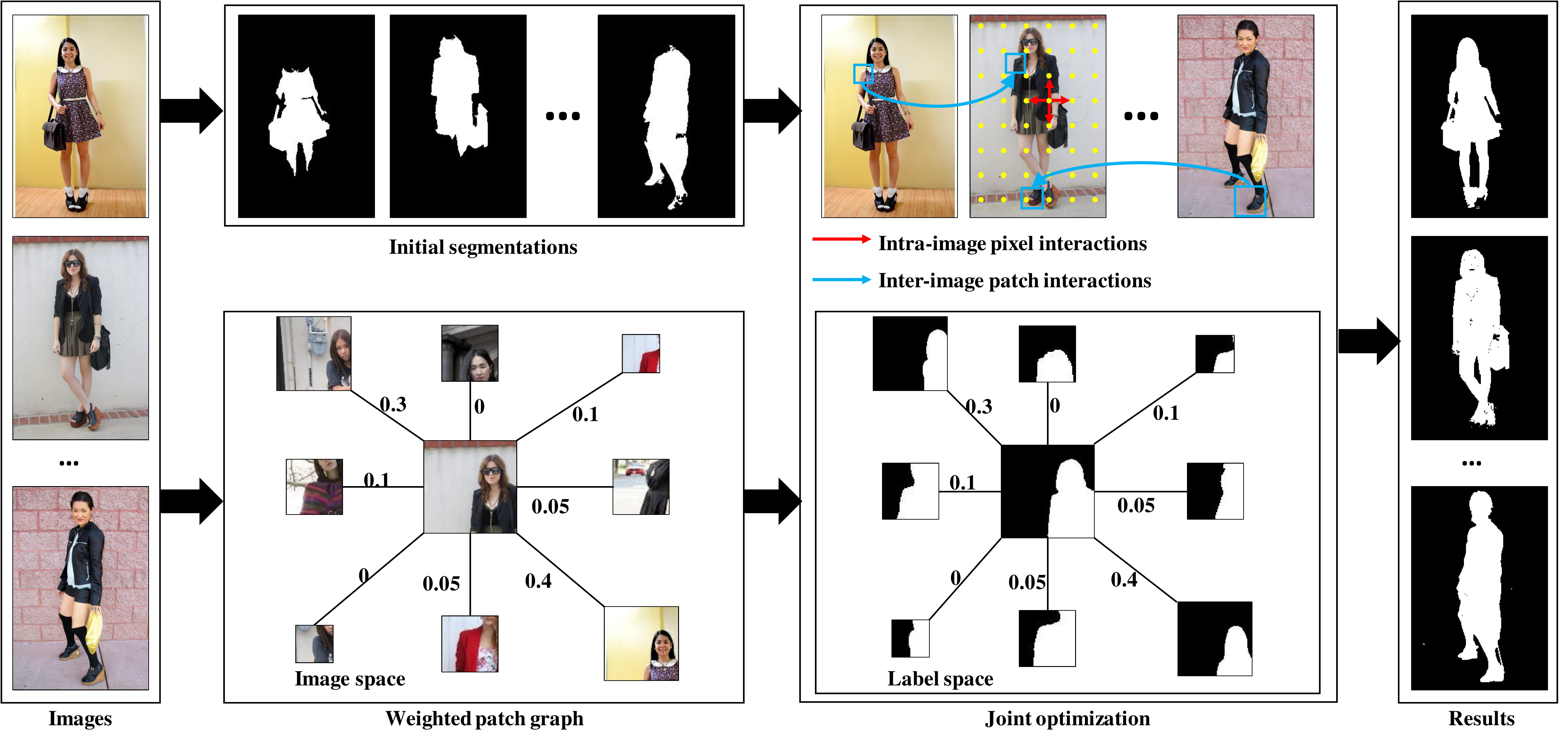}
	\caption{Framework of our approach. The initial foreground/background segmentation for each input image is estimated using~\cite{MB}. Meanwhile, we construct a weighted graph among the patches sampled from different images using~\cite{matchingCVPR2016}, where weights are learned by Locally Linear Embedding~\cite{lle}. Finally, we optimize intra-image object segmentation and inter-image local shape transfer jointly while preserving the patch weights in label space.}
	\label{fig:overview}
\end{figure*}
\subsection{Matching-based Group}
Matching-based approaches focus on building the region correspondences among different images. Some works mainly examine the matching constraint at object-level, and assumed that foreground feature histograms aggregated on different images are similar~\cite{RotherCVPR06,short_path-seg,short_path_2013,WangICCV13}. However, this strategy may have difficulty in applying to object categories with large variation. 
Another idea is conducting image co-segmentation through local region correspondences among images.
For example, Wang \etal~\cite{fmaps} proposed to match regions in the functional space. Rubio \etal~\cite{Unsupervised} developed a MRF formulation to jointly address object co-segmentation and region matching. 
Yu \etal~\cite{YUICME14} explored the inter-image similarity using a simple superpixel matching algorithm.
After obtaining the correspondences among those superpixels, they transferred the foreground/background labels among the matched pixels/superpixels. 
Faktor and Irani~\cite{FaktorICCV13} adopted structured matching to detect the common object parts in different images, through which ``co-occurring'' maps were generated to guide segmentation in each image. 
Lee \etal~\cite{co_cvpr15}suggested a multiple random walkers (MRW) clustering approach to extract the common objects from image set.
More recently, with the rise of visual saliency (\eg~\cite{xilin_icip13, yongdong_icme14,shiguang_cvpr14, MB,jia_ijcv,saliencyCVPR2016}) in computer vision field, some studies adopted visual saliency in image co-segmentation. 
For example,
Jerripothula \etal~\cite{co-fusion} proposed a saliency co-fusion-based co-segmentation method.
Liu \etal~\cite{pami_co17} employed co-saliency maps of an image set to guide clustering  image elements (\ie, superpixels) into two classes.
What is more, Wang \etal~\cite{tip_co17} calculated co-occurrence maps of the common objects for image cosegmentation.

Our approach is also based on local region matching. 
Different from Rubio~\etal~\cite{Unsupervised} and Wang \etal~\cite{fmaps}, we transfer labels at patch-level rather than point-level. In this manner, structured consistency is imposed to preserve the local object shapes during transfer. Compared with~\cite{co-fusion}, our approach does not assume the ``co-saliency'' of the common objects in the whole image set. In contrast, we assume only the co-occurrence of the parts of a common object in a sparse set of neighboring image patches. As a result, the proposed approach can robustly and  effectively identify the whole foreground objects, as confirmed by extensive experiments.

\subsection{Shape Transfer}
Shape transfer is a young yet widely adopted approach for data-driven foreground/background segmentation. Most existing works transfer the masks of pre-segmented objects to the test images, \eg,~\cite{TigheCVPR13} and \cite{KuettelCVPR12}. Beyond global object shapes, recent studies~\cite{XiaTCSVT15},~\cite{patch_cut} pursed to adopt local shape masks for image segmentation. 
Xia~\etal~\cite{XiaTCSVT15} proposed to infer foreground objects masks through sparse representations over global objects masks and local patch-level masks of the training set.
Yang~\etal~\cite{patch_cut} adopted dense correspondences among images to find candidate local
shape masks for each patch of a test image in an example image set, then they investigated an object segmentation scheme by patch-level local shape transfer.
Notably, local patch strategy help overcome the local deformations, which be proved by~\cite{XiaTCSVT15},~\cite{patch_cut}. Our work is inspired by local shape transfer successes, but we operate image co-segmentation in an unsupervised method without assuming pre-segmented images at hand.

\section{Co-segmentation Framework}
\subsection{Overview}
The procedure of our approach is shown in Fig.~\ref{fig:overview}.
Given a set of $M$ images, our framework first estimates a coarse initial foreground segmentations by thresholding saliency maps~\cite{MB}, which provides known cues for learning Gaussian Mixture Models~(GMM) in the following optimization. Meanwhile, we
generate a number of patches on each image by multi-scale sliding windows and then 
 construct a weighted patch graph to implement the transfer of patch-level local shape across images. Then to refine the segmentations in all images jointly, we integrate the patch graph and the coarse initial segmentation into a uniform framework.
 In the rest of this section, we first explain how local shape transfer helps image co-segmentation and the patch graph implementation, and then we describe our co-segmentation algorithm in detail.
 
 \begin{figure*}
 	\centering
 	\includegraphics[width=0.98\textwidth]{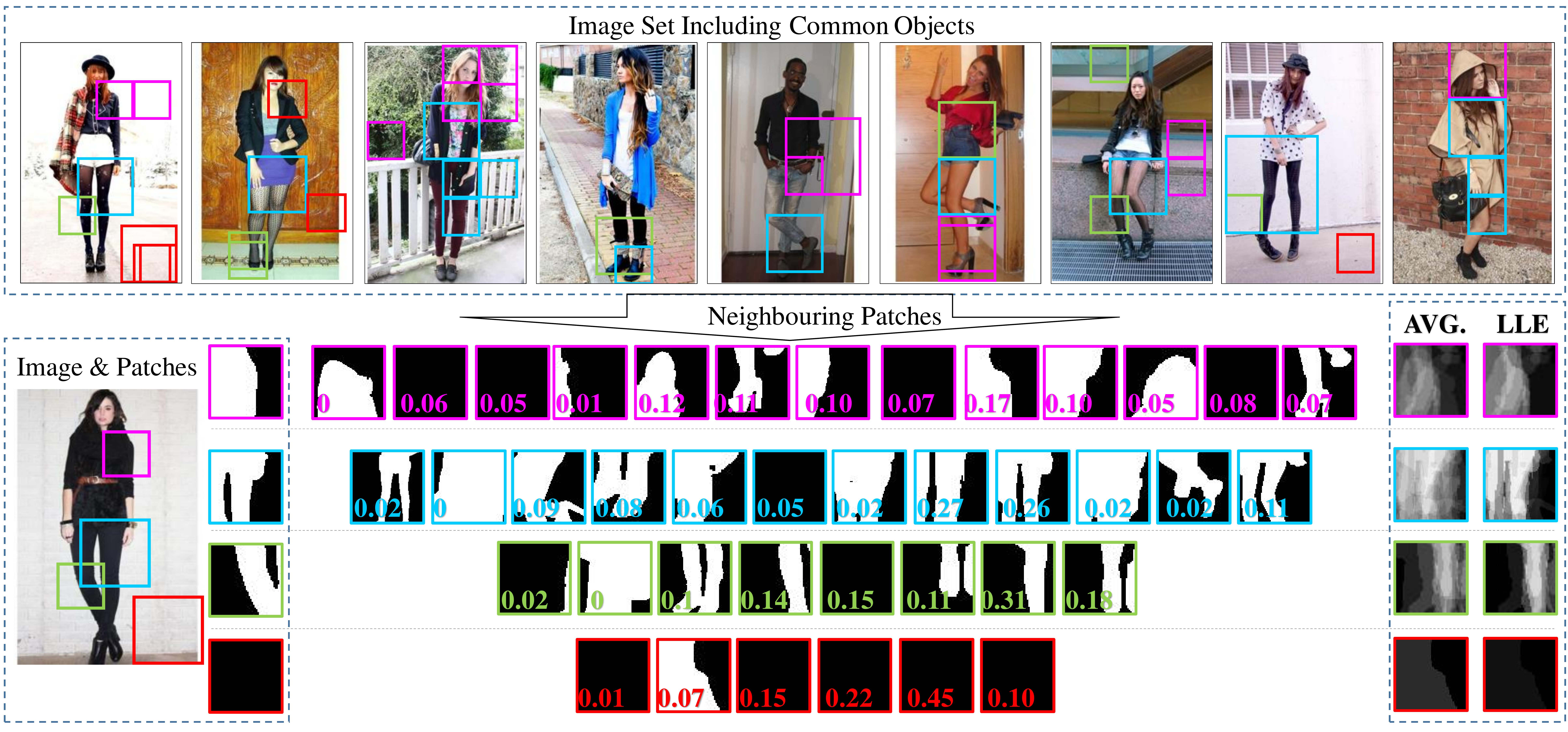}
 	\caption{Illustration of local shape transfer. For the patches sampled from an image (bottom-left), we illustrate the neighbouring patches~(bottom-middle) on different images. Different colors represent different patches and their neighbours. The transferred segmentation mask using average pooling~(AVE.) and Locally Linear Embedding~(LLE) are shown on the bottom-right. The learned weights at the bottom-left of each neighbouring patch show that Locally Linear Embedding is effective to suppress incorrect matches, which leads to more reliable shape transfer results.}
 	\label{fig_lle}
 \end{figure*}
\subsection{Local Shape Transfer}
From the machine learning perspective, the holistic shape of common objects in real-world lies in a high-dimensional space. In existing studies, object shape spaces are often learned from sophisticated non-linear models (\eg,~random forests~\cite{LiuTIP15}). However, our observation finds that the local object shapes can be well represented by their sparse neighbours in a linear and low-dimensional space. To this end, we first sample a number of patches for each image by multi-scale sliding windows. Then to connect the patches among different images, we construct a neighbourhood system through finding patch-level correspondences across images~\cite{matchingCVPR2016}. Specifically, let $p_1$ represent the collection of patches from image ${I_1}$ and $p_2$ denote the patch set from image $p_2$.
For each patch in $p_1$, the algorithm in~\cite{matchingCVPR2016} calculated matching patch in $p_2$ and vice versa. 
Throughout the whole image set, ${M-1}$ matching operations are employed for image ${I_1}$.With this manner, for the $i$th patch in a set of M images, we readily obtain its neighbours from different images, and represent the indices of these neighbours with $\mathcal{N}_i$. 

To represent $i$th patch with its neighbouring patches, 
we normalize all patches to a uniform size of $48 \times 48$.
Then one straightforward approach assumes that the neighbouring patches are all good surrogates of the original image patch. Based on this assumption, this approach formularizes the segmentation of the $i$th patch by the average of its neighbours, that is, 
$\vec y_i \approx \frac{1}{{\left| {{{\cal N}_i}} \right|}}\sum\limits_{j \in {{\cal N}_i}} {\vec y}_j$. In this formula, $\vec{y}_i$ denotes the concatenation of binary segmentation labels in the $i$th patch and ${{\left| {{{\cal N}_i}} \right|}}$ represents the number of neighbours . However, constructing primary patch by this formulation is difficult in real-word. As shown in the right column of Fig.~\ref{fig_lle},
aggregating the neighbouring patches with the inconsistent structure may confuse the shape transfer. To address this issue, we apply Locally Linear Embedding~\cite{lle} to learn a sparser but more reliable neighbouring relationships for each patch:
\begin{equation}
\min_{\bf{w} \succeq 0} \sum\limits_{i = 1}^P {\left\| {{\vec{x}_i} - \sum\limits_{j \in \mathcal{N}_i} {{w_{ij}}{\vec{x}_{j}}} } \right\|} ^2, \
s.t. \ \forall i, \ \sum\limits_{j \in \mathcal{N}_i} {{w_{ij}}}  = 1,
\label{eq:lle}
\end{equation}
where $P$ is the total number of patches sampled from the image set, and $\vec{x}_i$ is the Histograms-of-Oriented~(HOG) feature~\cite{hog_feature} extracted from the $i$th patch. The simplex constraint imposes sparsity on neighbour selection. Given a set of weights $w_{ij}$ and neighbors $\mathcal{N}_i$,
 the local shapes are thus transferred by $\vec{y}_i \approx \sum_{j \in \mathcal{N}_i} w_{ij} \vec{y}_j$. Fig.~\ref{fig_lle} illustrates that this strategy leads to more consistent shape transfer results.

\subsection{Image Co-segmentation by Local Shape Transfer}
Given the patch graph, the initial segmentation in each image can be refined by transferring the multi-scale local shapes from other images in the set. During the transfer, the weights learned in the patch feature
space are preserved for optimizing the label space. The objective of our algorithm is written formally as follows:
\begin{equation}
\begin{split}
&\min_{\bf{y}} \sum_{i=1}^M E_{\rm{seg}}({\bf{y}}^{[i]}) 
 + {\alpha} {\sum\limits_{i = 1}^P {\left\| {{\vec{y}_i} - \sum\limits_{j \in {N_i}} {{w_{ij}}{\vec{y}_{j}}} } \right\|} ^2}, \\
& s.t. \ \bf{y} \in \{\rm{0}, 1\}^{|\bf{y}|},
\label{eq:form}
\end{split}
\end{equation}
where $\bf{y}$ concatenates the foreground/background labels of all pixels in the image set, ${\bf{y}}^{[i]}$ is the binary segmentation of $i$th image and $\vec{y}_{i}$ denotes the foreground/background labels of $i$th patch. The energy term $E_{\rm{seg}}$ implements intra-image foreground/background segmentation, for which we use the popular Markov Random Field (MRF) energy~(see~\cite{BoykovPAMI11} for details). The problem~\eqref{eq:form} is NP-hard and usually a large scale as it operates on pixels. In order to seek the 	
optimum solution of~\eqref{eq:form}, 
we propose an efficient algorithm to approximate it by half quadratic splitting~\cite{Geman1992}.

The core idea of our optimization algorithm is introducing an auxiliary
variable $\bf{z}$ using $\vec{z}_i = \vec{y}_i$ as a constraint. 
Then Equation~\eqref{eq:form} can be relaxed to:
\begin{equation}
\begin{split}
&\min_{{\bf{y}}, {\bf{z}}} \sum_{i=1}^M E_{\rm{seg}}({\bf{y}}^{[i]}) 
+ {\alpha} {\sum\limits_{i = 1}^P {\left\| {{\vec{z}_i} - \sum\limits_{j \in {N_i}} {{w_{ij}}{\vec{z}_{j}}} } \right\|} ^2} \\
&~~~~~~~~~~~~~~~~~~~~~~+ \lambda \sum_{i=1}^P \left\| \vec{z}_i - \vec{y}_i \right\|^2, \\
& s.t. \ {\bf{y}}, {\bf{z}} \in \{\rm{0}, 1\}^{|\bf{y}|}.
\label{eq:reform}
\end{split}
\end{equation}

To find optimal solution for $\bf{y}$ and $\bf{z}$, we adopt an iterative process approach, in which $\bf{y}$ and $\bf{z}$ are optimized though keeping one of the them fixed. With this manner, the original problem is decoupled into two simple sub-problems:
\begin{equation}
\begin{split}
&\min_{{\bf{y}}} \sum_{i=1}^M E_{\rm{seg}}({\bf{y}}^{[i]}) + \lambda \sum_{i=1}^P \left( \left\| \vec{y}_i \right\|^2 - 2 (\vec{z}_i)^{\mathrm{T}} \vec{y}_i \right), 
\\
& s.t. \ \bf{y} \in \{\rm{0}, 1\}^{|\bf{y}|}.
\end{split}
\label{eq:solve_y}
\end{equation}
\begin{equation}
\begin{split}
&\min_{{\bf{z}}}  {\alpha} {\sum\limits_{i = 1}^P {\left\| {{\vec{z}_i} - \sum\limits_{j \in {N_i}} {{w_{ij}}{\vec{z}_{j}}} } \right\|} ^2}
 + \lambda \sum_{i=1}^P \left\| \vec{z}_i - \vec{y}_i \right\|^2, \\
&s.t. \ {\bf{z}} \in \{\rm{0}, 1\}^{|\bf{y}|}.
\label{eq:solve_z}
\end{split}
\end{equation}

In formula~\eqref{eq:solve_y}, patch label $\left\| \vec{y}_i \right\|^2=\left\| \vec{y}_i \right\|$ is binary. When $\bf{z}$ is fixed, we change the second term of~\eqref{eq:solve_y} into a linear form of~\wrt~$\bf{y}$, and then directly merge it into the unary potentials of the MRF energy. Consequently, by performing graph-cut~\cite{GrabCut} in each image, 
this
new expression of~\eqref{eq:solve_y} is efficient.
The formula~\eqref{eq:solve_z} aims to optimize~$\bf{z}$ while keeping $\bf{y}$ fixed. To solve such a large-scale quadratic program, 
we first discard the
binary constraint of the variable $\bf{y}$.
Then by solving a
linear system, 
 we obtain a closed-form solution of the quadratic program~\eqref{eq:solve_z}. More specifically, we approximate this solution by a sequence of label diffusions.
In a diffusion step, the pixel label~$\vec{z}_i$ in the $i$th patch is optimized by fixing the labels of all other pixels. By setting the derivation \wrt~$\vec{z}_i$ as zero, we obtain the following updated rule
\begin{equation}
\begin{split}
&{{\vec z}_{i'}} = \frac{{\alpha \left[ {\sum\limits_{j \in {{\cal N}_i}} {{w_{ij}}} {{\vec z}_j} + \sum\limits_{j:i \in {{\cal N}_j}} {{w_{ji}}} \left( {{{\vec z}_{j\_i}}} \right)} \right] + \lambda {{\vec y}_i}}}{{\alpha  + \lambda  + \sum\limits_{j:i \in {{\cal N}_j}} {w_{ji}^2} }},\\
&{{\vec z}_{j\_i}} = {{\vec z}_j} - \sum\limits_{k \in {{\cal N}_j},k \ne i} {{w_{jk}}} {{\vec z}_k}.
\end{split}
\end{equation}
For easy presentation, this formula can be written in a compact form: ${{\bf{Z}}^\prime } = AB$. Here ${A}$ and ${B}$ are defined as:
\begin{equation}
\begin{split}
&A = \lambda {\bf{Y}} + \alpha \left[ {{\bf{W}} + {{\bf{W}}^{\rm{T}}} - {{\bf{W}}^{\rm{T}}}{\bf{W}} + {\rm{Diag}}({{\bf{W}}^{\rm{T}}}{\bf{W}})} \right]{\bf{Z}},\\
&B = {\left[ {(\alpha  + \lambda ){\bf{I}} + {\rm{Diag}}({{\bf{W}}^{\rm{T}}}{\bf{W}})} \right]^{ - 1}},
\end{split}
\end{equation}
where the matrices ${\bf{Z}}$, ${\bf{Z}}'$ and ${\bf{Y}}$ concatenate in a row of the column vectors $\vec{z}_i$, $\vec{z}_i'$ and $\vec{y}_i$, respectively, ${\bf{W}}$ is a $P \times P$ pairwise matrix of patch-wise neighbouring weights, and ${\bf{I}}$ is the identical matrix. The operator $\mathrm{Diag}~(\cdot)$ creates a diagonal matrix by picking out the diagonal elements of the input matrix. In practice, after a diffusion step, we normalize the soft labels ${\bf{z}}$ into $[0,1]$ for each image. Finally, we terminate the diffusion step when previous and current $\vec{z}_i$ are less than the threshold values set manually.

The two steps are repeated until near-convergence. Empirically, we terminate the optimization in 10 iterations and take the last discrete labels y as the final segmentations.

\subsection{Comparison with our previous method~\cite{dlle}}
The main difference lies in that this work incorporates multi-scale strategy into image cosegmentaion. In particular, we first use multi-scale windows to sample images instead of fixed the patch size like previous work~\cite{dlle}. Then we directly find the neighboring patches on patch-level matching rather than on pixel-level dense correspondences, which guarantees the computational efficiency of our approach.

\begin{table*}
	\centering
	\begin{threeparttable}
		\caption{\label{tab:subicoseg_acc}Comparison with leading co-segmentation approaches of
			correctly classified pixels (denoted by ACC) on iCoseg dataset. }
		\begin{tabular}{cccccccccccc}
			\toprule
			iCoseg &Ours &DLLE\cite{dlle} &Lee~\cite{co_cvpr15} &Fu~\cite{Object-based} &Wang~\cite{fmaps} &Liu~\cite{pami_co17} &Rubio~\cite{Unsupervised} &Vicente~\cite{Object}&Mukherjee~\cite{MukherjeeECCV12}&Joulin~\cite{discriminative}\\
			\midrule
			Alaska Bear &0.926 & 0.861 & 0.873 & \underline{\textbf{0.935}} & 0.904 &0.872 & 0.864 & 0.900 & - & 0.748\\
			Red Sox Players &0.964 & \underline{\textbf{0.972}} & 0.971 & 0.965 & 0.942 &0.927 & 0.905 & 0.909 & 0.957 & 0.730\\
			Stonehenge1 &0.938  & 0.936 & \underline{\textbf{0.959}} & 0.930 & 0.925 &0.820 & 0.873 & 0.633 & 0.927 & 0.566 \\
			Stonehenge2 &\underline{\textbf{0.946}}  & 0.844 & 0.907 & 0.835 & 0.872 &0.800 & 0.884 & 0.888 & 0.849 & 0.860 \\
			Liverpool &0.902  & 0.905 & 0.885 & \underline{\textbf{0.921}} & 0.894 &0.911 & 0.826 & 0.875 & - & 0.764 \\
			Ferrari &0.874  & 0.892 & 0.919& 0.917 & \underline{\textbf{0.956}} &0.900 & 0.843 & 0.899 & 0.900 & 0.850 \\
			Taj Mahal &0.810   & 0.878 & \underline{\textbf{0.952}} & 0.887 & 0.926 &0.832 & 0.887 & 0.911 & 0.941 & 0.737\\
			Elephants &\underline{\textbf{0.974}}  & 0.961 & 0.931 & 0.904 & 0.867 &0.900 & 0.750 & 0.431 & 0.877 & 0.701\\
			Pandas &0.927  & 0.835 & 0.848& 0.812 & 0.886 &0.800 & 0.600 & 0.927 & \underline{\textbf{0.928}} & 0.840\\
			Kite &\underline{\textbf{0.984}}  & 0.980  & 0.957 & 0.966 & 0.939 &0.978 & 0.898 & 0.903 & 0.946 & 0.870\\
			Kite panda &0.944  & 0.905 & \underline{\textbf{0.960}} & 0.838 & 0.931 &0.812 & 0.783 & 0.902 & 0.934 & 0.732\\
			Gymnastics &\underline{\textbf{0.992}}  & 0.984 & 0.961 & 0.954 & 0.904 &0.969& 0.871 & 0.917 & 0.922 & 0.909\\
			Skating &0.924  & 0.893 & 0.916 & 0.817 & 0.787 &0.822 & 0.768 & 0.775 & \underline{\textbf{0.966}} & 0.821\\
			Hot Balloons &\underline{\textbf{0.991}}  & 0.969 & 0.977 & 0.965 & 0.904 &0.938 & 0.890 & 0.901 & 0.952 & 0.852\\
			Liberty Statue &\underline{\textbf{0.993}}  & 0.966 & 0.945 & 0.927 & 0.968 &0.957 & 0.916 & 0.938 & 0.966 & 0.906\\
			Brown bear &\underline{\textbf{0.962}}  & 0.938 & 0.937 & 0.948 & 0.881 &0.823 & 0.804 & 0.953 & 0.885 & 0.740\\
			Average &\underline{\textbf{0.941}}  & 0.920 & 0.931 & 0.907 & 0.905 &0.879 & 0.839 & 0.853 & - & 0.789\\
			\bottomrule
		\end{tabular}
		\small
	\end{threeparttable}
\end{table*}

\section{Experiments}
\subsection{Experimental Settings}
We first sample many patches on each image by four scales sliding windows, $48 \times 48$, $72 \times 72$, $96 \times 96$, $120 \times 120$. On each scale, there are no overlapping patches expect image boundary, which ensure that patches can cover the whole image.
 The unary term of the MRF energy $E_{seg}$ is the (log-negative) foreground/background color likelihoods generated by 12-components GMM models. Initially, the GMMs are learned on saliency-based segmentations. In each iteration, we update them using the latest segmentations. We follow~\cite{CasacaCVPR14} to define the pairwise term, modeling color contrasts between the adjacent pixels.
Parameters $\alpha$ and $\lambda$ are empirically set as $1$ and $0.3$, respectively.
We evaluate the proposed approach on three public benchmarks:
iCoseg dataset~\cite{icoseg}, MSRC dataset~\cite{MSRC}, Coseg-Rep dataset~\cite{cosketch}  and Fashionista dataset~\cite{parsing_clothing}. 
For easy of presentation,
we use MCO to represent the  proposed approach of multi-scale local shape transfer and SCO to denote a variant of turning the proposed approach into single-scale with patch size of $48 \times 48$. We also refer to the previous algorithm~\cite{dlle} as DLLE which utilizes the pixel-level dense correspondences among different images.

The iCoseg dataset~\cite{icoseg} contains 643 images of 38 object classes with pixel-level annotations. In each class, common objects have similar color but various locations and scales. We test a subset of 16 classes which are widely used by the leading co-segmentation approaches, and we also make a comparison with state-of-the-arts on the whole dataset of 38 classes. For each class, all images are used for our co-segmentation framework. 
The MSRC dataset~\cite{MSRC} contains 418 images from 14 categories. The objects in each class have various color, pose and scale, which adds difficulties to conduct co-segmentation. 
Coseg-Rep dataset~\cite{cosketch} has 23 categories with 572 images, where categories are different species of flowers and animals. Interestingly, there is a special category named "Repetitives" that each image from this category has several objects of similar shape patterns.
In our experiments, images from same class are segmented at once. 
The Fashionista dataset~\cite{parsing_clothing} contains 685 street photographs of fashion models. In contrast to conventional co-segmentation datasets, Fashionista dataset is extremely challenging with various human poses, background clutters and complex appearances. As existing co-segmentation approaches may have difficulty in operating large amounts of images, we randomly partition the dataset into 23 groups with nearly 30 images per group and resize them with a resolution of $300 \times 200$. Evaluations are averaged over 10 random partitions.

We use two evaluation protocols: the accuracy of correctly classified pixels (denoted by ACC) and intersection-over-union score (denoted by IOU).
Both agreements are chosen for throughout comparison with previous approaches, and the latter is more preferred as it has been shown unbiased to the object size~\cite{LiCVPR13}. 
Note that higher values of both accuracies means the better co-segmentation results. 
\subsection{Comparison with State-of-the-Arts}

\begin{table}
	\centering
	\begin{threeparttable}
		\caption{\label{tab:icoseg_acc}Comparison results of the proposed method and the state-of-the-art co-segmentation methods on the iCoseg dataset in terms of average ACC and IOU.}
		\begin{tabular}{ccccc}
			\toprule
			&\multicolumn{2}{c}{~~~~~sub-iCoseg~(16)~~~~~}&\multicolumn{2}{c}{~~~~~iCoseg~(38)~~~~~}\\
			&~~~~ACC &IOU &~~ACC &IOU\\
			\midrule      
			Ours  &~~~~0.941 &0.79 &~~0.925 &0.73\\
			Quan~\cite{QUANCVPR2016}  &~~~~0.948 &0.82 &~~0.933 &0.76 \\
			Faktor~\cite{FaktorICCV13}     &~~~~0.944 &0.79 &~~0.928 &0.73 \\
			Jerripothula~\cite{co-fusion}     &~~~~- &- &~~0.919 &0.72\\
			Kuettel~ \cite{segmentation_imagenet} &~~~~- &- &~~0.914 &-\\
			Dai~\cite{cosketch} &~~~~- &- &~~0.895 &- \\ 
			Meng~\cite{cviu_co16} &~~~~- &- &~~- &0.71 \\
			\bottomrule
		\end{tabular}
	\end{threeparttable}
\end{table}

\begin{figure*}
	\includegraphics[width=0.98\textwidth]{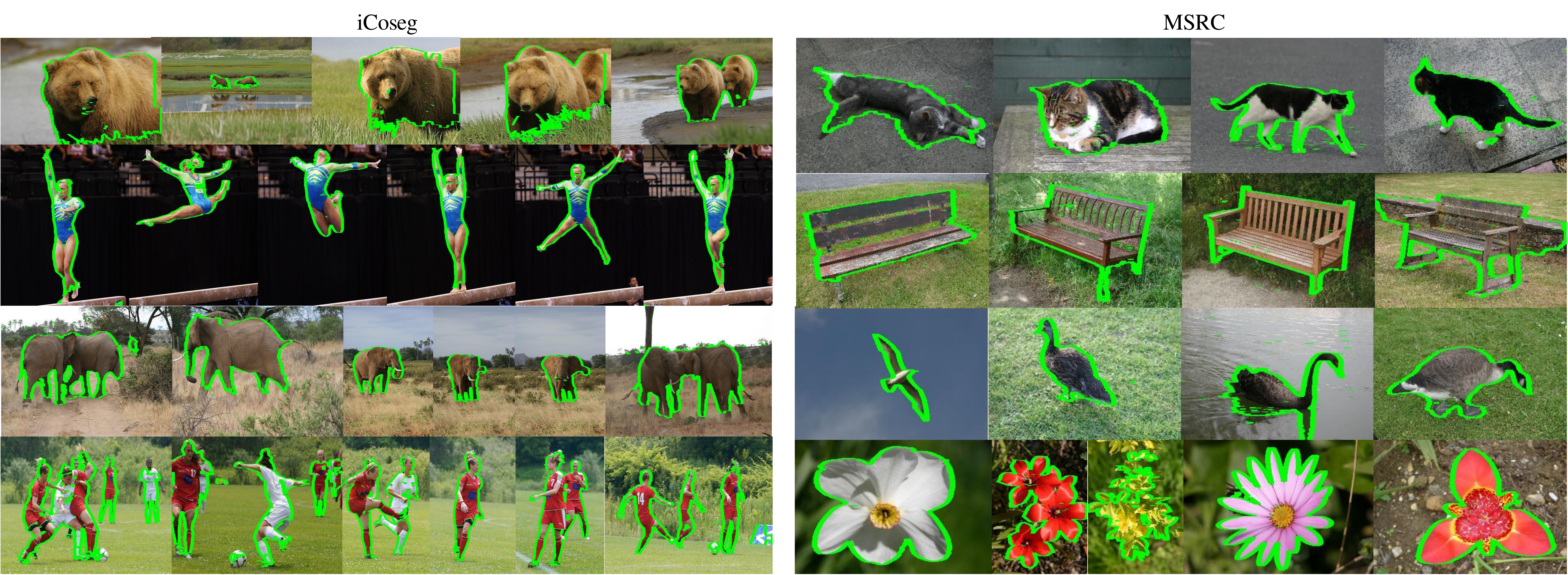}
	\centering
		\caption{Representative segmentations of our approach on iCoseg dataset~\cite{icoseg} and MSRC dataset~\cite{MSRC}. Note that foreground objects are marked by green lines.}
	\label{result_icoseg}
 \end{figure*}

\textbf{Comparisons on iCoseg Dataset}.
As common objects in each class of iCoseg are known to have similar color properties, we follow the suggestion of ‘Joint-Grab-Cut’~\cite{segmentation_imagenet}. Namely, we apply jointly updating the color models to all images rather than performing Grab-cut to each image separately.
we summarize the co-segmentation accuracies in Table1 and Table2.
In Table~\ref{tab:subicoseg_acc}, our approach outperforms other approaches based on local region matching~\cite{Unsupervised,fmaps}. We believe that it is the patch-level structured consistency that makes the difference. We also obtain better results than~\cite{Object,Object-based}, although they used external training data. Note that~\cite{MukherjeeECCV12} performs quite well on the reported 14 classes, achieving 92.49\% average accuracy, while our approach obtains 94.5\%. However, they also rely on training images to learn dictionaries while our approach is unsupervised. 
More specifically, our approach achieves the best overall performance with leading accuracies on $7/16$  categories. We obtain remarkable results on challenging categories, such as \textit{elephants}, \textit{stonegenge2} and \textit{gymnastics}. Although most previous approaches work less well as a result of these object categories with large pose variance, our proposed multi-scale local shape transfer strategy may handle them better. 
Interestingly, our multi-scale method outperforms our previous work~\cite{dlle} in 12 out of 16 categories.
In Fig.~\ref{result_icoseg}, we illustrate some related visual results of our co-segmentation approach. 
In Table~\ref{tab:icoseg_acc}, we show the comparison results of our method
and state-of-the-arts co-segmentation algorithms on a subset of the iCoseg
dataset (listed in Table~\ref{tab:subicoseg_acc}) as well as the whole iCoseg dataset. In terms of ACC from table~\ref{tab:icoseg_acc},
our approach performs better than~\cite{segmentation_imagenet, co-fusion} and comparably with~\cite{FaktorICCV13, QUANCVPR2016}. Although \cite{QUANCVPR2016} has reported the best performance so far on iCoseg dataset as result of using high-level semantic
features, all of~\cite{QUANCVPR2016, FaktorICCV13} and our approach can effectively locate the common objects on this dataset. And the main differences among these approaches are mainly due to finer localization of object boundaries. 

\begin{table}
	\caption{\label{tab:msrc_fa}Comparison results of the proposed method and the state-of-the-art co-segmentation methods on MSRC dataset and Fashionista dataset in terms of average ACC and IOU.}
	\centering
	\subtable[MSRC]{
		\begin{tabular}{ccc}
			\toprule
			&ACC&IOU\\  
			\midrule
			Ours &0.884 &0.70 \\
			Wang~\cite{tip_co17}  &0.909 &0.73 \\
			Faktor~\cite{FaktorICCV13} &0.892 &0.73 \\
			Jerripothula~\cite{co-fusion} &0.887 &0.71 \\
			Jerripothula~\cite{Meng_ICIP2015} &0.884 &0.70 \\
			Rubinstein~\cite{RubinsteinCVPR13} &0.877&0.68 \\
			\bottomrule
		\end{tabular}
		\label{tab:msrc_fa_msrc}
	}
	\subtable[Fashionista]{        
		\begin{tabular}{ccc}
			\toprule
			&ACC &IOU \\
			\midrule
			Ours &0.949&0.794\\
			DLLE~\cite{dlle} &0.937 &0.755 \\
			Faktor~\cite{FaktorICCV13} &0.869&0.501 \\
			Dai~\cite{cosketch} &0.862&0.576\\
			Joulin~\cite{discriminative} &0.724 &0.358 \\
			Rother~\cite{GrabCut} &- &0.642 \\
			\bottomrule
		\end{tabular}
		\label{tab:msrc_fa_fa}
	}
\end{table}

\textbf{Comparisons on MSRC Dataset}.
Unlike iCoseg dataset, common objects from MSRC dataset always have more variances in appearance.
And unfortunately, several images cannot obtain initial coarse masks by our previous simple thresholding strategy. Therefore, for each category in MSRC dataset, we apply saliency-cut~\cite{saliency-cut} to threshold the saliency maps. 
Then with these initial coarse masks, we conduct our co-segmentation approach on each category and summarize the results in Table~\ref{tab:msrc_fa_msrc} and Fig.~\ref{result_icoseg}. 
We can see that our average performance is competitive to \cite{co-fusion} as well as \cite{Meng_ICIP2015}. 
We also find that our method performs worse than \cite{FaktorICCV13},~\cite{tip_co17} on MSRC dataset. The main reason could be attributed to the dependence on saliency maps. If the saliency detection method cannot treat the common object as salient or turn background pixels into salient, our method would not be able to segment it out. 
Inspired by recent deep learning technology, we will try to integrate convolutional neural network into co-segmentation in future research.

\begin{table}[t]
	\centering
	\begin{threeparttable}
		\caption{\label{tab:coseg_rep}Comparison results of the proposed method and the state-of-the-art co-segmentation methods on Coseg-Rep dataset in terms of average ACC and IOU.}
		\begin{tabular}{ccccc}
			\toprule
			&Ours  &Dai~\cite{cosketch}  &Jerripothula~\cite{Meng_ICIP2015} &Jerripothula~\cite{co-fusion}\\
			\midrule
			ACC &0.932 &0.902          &0.922                      &0.934\\
			IOU &0.78  &0.67            &0.73                        &0.77\\
			\bottomrule
		\end{tabular}
	\end{threeparttable}
\end{table}

\textbf{Comparisons on Coseg-Rep Dataset}.
This dataset has 23 object categories and 572 images in total.  Among them, a special category called
"repetitive" includes a variety of animals and flowers. To conduct our approach on this special category, we first divide the "repetitive" category into two subcategory (one subcategory only contains animals, the other subcategory only includes flowers). Then to avoid that several images cannot obtain coarse masks by simple thresholding saliency maps, we apply saliency-cut~\cite{saliency-cut} to threshold the saliency maps. After that, we conduct our co-segmentation method on each category or subcategory, and summarize the results in Table~\ref{tab:coseg_rep}.  In terms of intersection-over-union (denoted by IOU) score, it can be seen from Table~\ref{tab:coseg_rep} that our approach outperforms the state-of-the-art methods on Coseg-Rep dataset. Moreover, even though~\cite{co-fusion} tuned its parameter over categories,
our method achieve 1\% improvement of IOU score when compared with the best results reported in~\cite{co-fusion}.

\begin{figure*}
	\centering
	\includegraphics[width=0.98\textwidth]{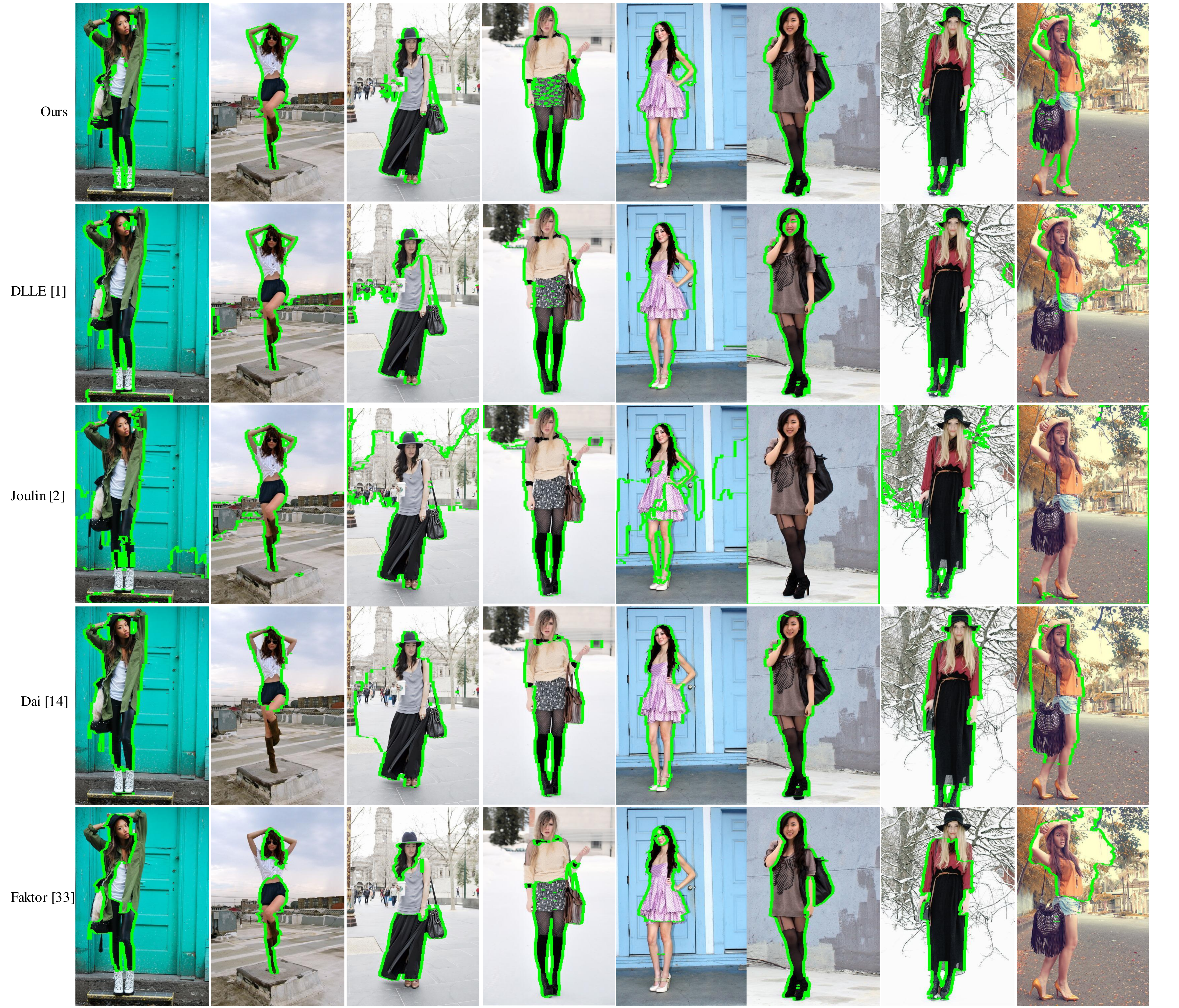}
	\caption{Visual
		comparisons between our method and state-of-the-arts. The segmentation results of \cite{discriminative},\cite{cosketch}, \cite{FaktorICCV13} are obtained by their released code. 
		And foreground objects are surrounded by green lines.}
	\label{result_fa}
\end{figure*}

\textbf{Comparisons on Fashionista Dataset}.
To further prove the effectiveness of our approach and clarify our contributions, we apply state-of-the-arts~\cite{FaktorICCV13,cosketch,discriminative} methods on the Fashionista dataset using the released codes. We also compare with a GrabCut~\cite{GrabCut} baseline using a bounding box with 8 pixels margin from the image borders. We summarize the evaluations
in Table~\ref{tab:msrc_fa_fa}, where the values of~\cite{FaktorICCV13,cosketch, discriminative} and our approach are averaged on all groups, while the values of the GrabCut baseline is directly taken from~\cite{patch_cut}.
According to Table~\ref{tab:msrc_fa_fa}, we find that the leading co-segmentation approaches have difficulty in generalizing this dataset. Our approach obtains desirable performance on Fashionista dataset. Notably, we obtain 58\%, 38\% and 122\% relative improvements over~\cite{FaktorICCV13},~\cite{cosketch} and~\cite{discriminative} in terms of IOU score on this dataset, respectively. Due to the complexity and large variance of object appearance and pose, the template-based approaches~\cite{cosketch,discriminative} may have difficulty learning a proper template to represent the object category, while~\cite{FaktorICCV13} often detects incomplete object shapes and misses important object details. On the contrary, the proposed multi-scale local shape transfer can successfully deal with the appearance and pose variances on Fashionista dataset. In Fig.~\ref{result_fa}, we give some visual comparisons between our algorithm and these well-known cosegmentation methods:~\cite{cosketch,discriminative,FaktorICCV13}. 

\textbf{Comparisons with Previous Work}.
Compared with our previous work DLLE~\cite{dlle}, the strategy of multi-scale local shape transfer is 2.1\% higher on sub-iCoseg dataset and 3.9\% higher on Fashionista datset. Significant improvements are observed on several categories~(\eg,\textit{Stonehenge2}, \textit{Kite Panda}, \textit{Pandas}), since the proposed multi-scale local shape transfer can better handle large variation of objects scales.
In order to clearly understand
the contributions of the new algorithms proposed, we turn the proposed approach~(MCO) into single-scale variant~(SCO) with patch size of $48 \times 48$ and conduct some additional experiments for MCO, SCO and DLLE~\cite{dlle}. Specifically,  
we select 30 images~($600 \times 400$) randomly from Fashionista dataset~\cite{parsing_clothing} and resize the images to $300 \times 200$. 
Without the loss of generality, we average IOU scores of ten groups experiments generated by MCO, SCO and DLLE~\cite{dlle}, respectively. Table~\ref{tab:com_hog} summarizes the average IOU results.

\begin{table}[t]
	\centering
	\begin{threeparttable}
		\caption{\label{tab:com_hog} Intersection-over-Union(IOU) scores on selected subset of Fashionista dataset}
		\begin{tabular}{cccc}
			\toprule
			&~~~~~~~MCO~~~~~~~&~~~~~~~SCO~~~~~~~&~~~~~~~DLLE~\cite{dlle}~~~~~~~\\
			\midrule
			images  &30 &30 &30\\
			IOU  &0.800 &0.777 &0.753 \\
			\bottomrule
		\end{tabular}
		\small
		Note: MCO is our proposed approach with multi-scale local shape transfer, SCO is a variant of turning the proposed approach into single-scale with patch size of $48 \times 48$, and DLLE refers to our previous algorithm~\cite{dlle}.
	\end{threeparttable}
\end{table}

Table~\ref{tab:com_hog} illustrates that SCO improves IOU over the previous algorithm DLLE~\cite{dlle}, which confirm the effectiveness of adopting proposal flow and HOG features to build the patch neighbourings system. Moreover, the proposed approach~(MCO) achieves improved performance than the variant method SCO, which confirm the effectiveness of adding multi-scale to the procedure of
local shape transfer.
With multi-scale local shape transfer
and patch neighbouring system, our proposed method MCO achieves substantial
improvement than our previous approach~\cite{dlle}. 
In Fig.~\ref{result_fa},
we provide some visual comparisons between our presented
approach~MCO and our previous work DLLE~\cite{dlle}. Obviously, the proposed
method MCO segment common objects more precisely than
previous DLLE, such as legs and arms.

\begin{table}[t]
	\centering
	\begin{threeparttable}
		\caption{\label{tab:alpha_test}Intersection-over-Union(IOU) scores of with or without multi-scale local shape transfer during segmenting on Fashionista dataset and sub-iCoseg dataset(including 16 classes listed in table~\ref{tab:subicoseg_acc}).}
		\begin{tabular}{cccc}
			\toprule
			&~~~~~dataset~~~~~& ~~~~~'w/o lst'~~~~~ & ~~~~~'ours'~~~~~\\
			\midrule
			Fashionista  &685 & 0.688 & 0.794\\
			sub-icoseg  &122 &0.548  & 0.790\\
			\bottomrule
		\end{tabular}
		\small
		Note:'w/o lst' and 'ours' denote the image co-segmentation results that without local shape transfer and our complete framework, respectively.
	\end{threeparttable}
\end{table}

\textbf{Running Time}.
Our approach takes around 5 minutes to process 30 images with resolution
$300 \times 200$. Saliency estimation can be done in a few seconds. Building correspondences,
learning graph weights and optimization take around 2.7, 0.02 and 2 minutes, respectively.
Empirical comparisons show that the current implementation runs much faster than several state-of-the-arts ~\cite{FaktorICCV13}~\cite{cosketch}. For example, Factor~\etal~\cite{FaktorICCV13} takes about 48 minutes and Dai~\etal~\cite{cosketch} takes about 58 minutes, using their released code.

\subsection{Performance Analysis}
In this section, we aim to study how the proposed approach works and further demonstrate
its effectiveness. To this end, we conduct two additional experiments. 

The first experiment conducts on Fashionista dataset~\cite{parsing_clothing} and sub-iCoseg dataset~\cite{icoseg} (including 16 classes listed in table~\ref{tab:subicoseg_acc}). The results desmonstrate the performance improvement of our algorithm after using multi-scale local shape transfer, and we summarize the results in table~\ref{tab:alpha_test}. Note that when no local shape transfer~('w/o lst') is employed, this variant can be seen as grab-cut with saliency cues directly. However, it is observed that our complete framework with multi-scale local shape transfer significantly improves the results, confirming the effectiveness of multi-scale local shape transfer.

\begin{table}[t]
	\centering
	\begin{threeparttable}
		\caption{\label{tab:multi-scale-test}Intersection-over-Union(IOU) scores on Fashionista dataset.}
		\begin{tabular}{ccc}
			\toprule
			&multi-scale patch size& ~~~~~~~~~~IOU~~~\\
			\midrule
			1~~~ & $16 \times 16$,~$40 \times 40$,~$64 \times 64$,~$88 \times 88$  &~~~~~~~~~~0.786~~ \\
			2~~~ & $32 \times 32$,~$56 \times 56$,~$80 \times 80$,~$104 \times 104$  &~~~~~~~~~~0.797~~ \\
			3~~~ &$48 \times 48$,~$72 \times 72$,~$96 \times 96$,~$120 \times 120$  &~~~~~~~~~~0.794~~ \\
			\bottomrule
		\end{tabular}
		\small
	\end{threeparttable}
\end{table}

In the two experiment, we investigate the segmentation accuracy as a function of the patch size of multi-scale. To this end, we adopt three different multi-scale strategies on Fashionista dataset~\cite{parsing_clothing} and summarize the performance in Table.~\ref{tab:multi-scale-test}. Without the loss of generality, we repeat this step for 10 times and each time randomly partition Fashionista dataset~\cite{parsing_clothing} into 23 groups with nearly 30 images per group. 
Table.~\ref{tab:multi-scale-test} shows average IOU scores of 10 experiments. Although the second strategy~($32 \times 32$,~$56 \times 56$,~$80 \times 80$,~$104 \times 104$) obtain the best performance on IOU score, the third strategy~($48 \times 48$,~$72 \times 72$,~$96 \times 96$,~$120 \times 120$) can achieve comparable IOU score. Note that IOU scores of the three different multi-scale strategies are close, this can be regarded as multi-scale strategy independent of patch size in certain degree. 
And we adopt the third strategy~($48 \times 48$,~$72 \times 72$,~$96 \times 96$,~$120 \times 120$) in our experiments.
Substantial comparison experiments illustrated in Table.~\ref{tab:subicoseg_acc}-\ref{tab:msrc_fa} show that our multi-scale strategy  performs efficiently on various benchmark datasets.

\begin{figure}[t]
	\centering
	\includegraphics[width=0.48\textwidth]{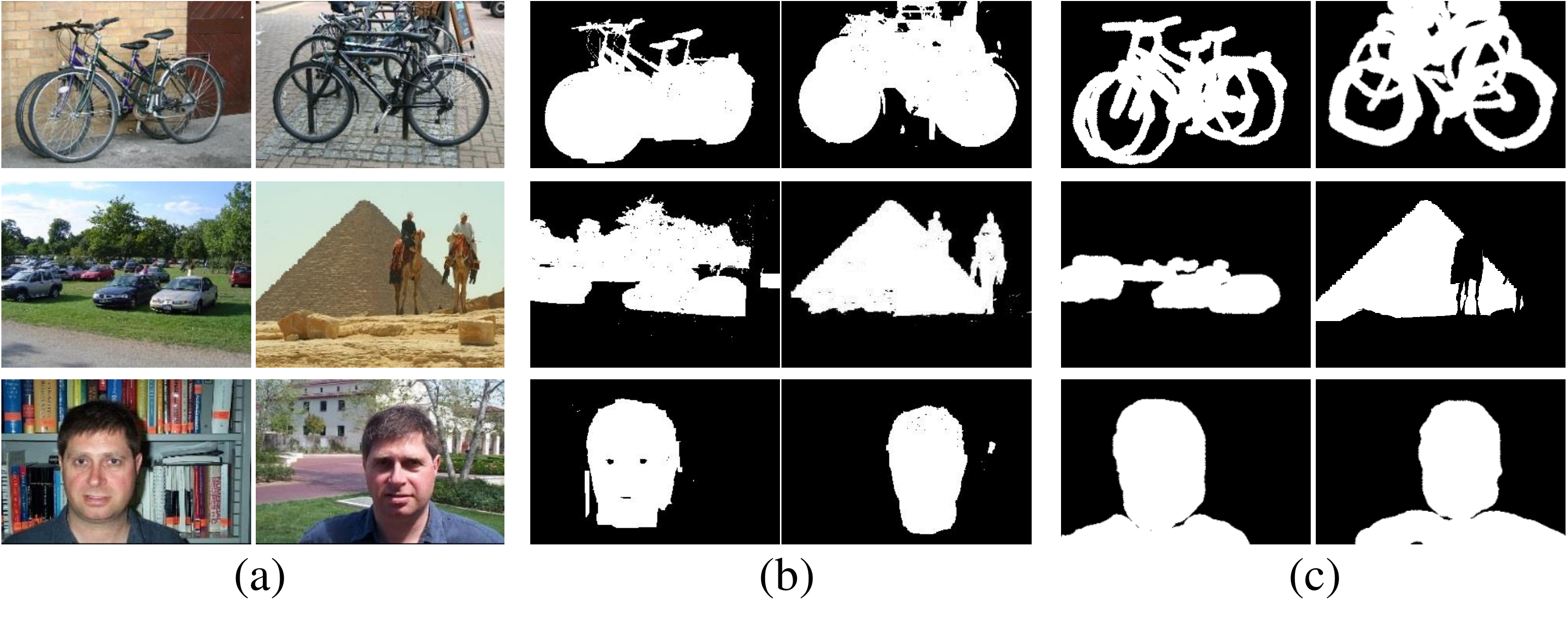}
	\caption{Failure cases. (a) The input images. (b) Co-segmentation results by our approach. (c) The ground truth masks.}
	\label{failure_img}
\end{figure}

\textbf{Failure cases}.
In Fig.~\ref{failure_img}, we show several typical failure cases of our approach. 
In the first row, our method can localize the bicycles whereas failing in segment thin rods and tires.
In the second row, examples show that the proposed approach fails to segment the target objects if images contain multiple common salient objects~(\eg tree and car).
In the third row, our approach cannot segment shoulders from background because shoulders are not salient in face category. These are common and challenge problems for image co-segmentation~(\eg~\cite{co-fusion}).

\section{Conclusion}
This study proposes an unsupervised approach of multi-scale local shape transfer for image co-segmentation. It starts with generating saliency maps and multi-scale patches on each image. Then we construct a reliable patch neighbourhood system and incorporate label consistencies among neighbouring patches in different images. Finally, the common objects are segmented through a graph-cut based algorithm that can generate binary mask for each image.
Extensive experiments demonstrate that the proposed algorithm of multi-scale local shape transfer can significantly boost the co-segmentation
performance. Compared with state-of-the-arts, our approach performs comparably on iCoseg dataset and MSRC dataset, and substantially better on Coseg-Rep dataset and the challenging Fashinista dataset.

Our results also reveal that local shape transfer among images is valuable for distinguishing the common foreground objects from complex background. We believe that precise local shape correspondences are a reliable way to handle image co-segmentation. In the future, we will further explore the usage of local shape transfer in image co-segmentation.  In particular, we will try some other weights learning methods (like popular deep learning architectures)
to build a more reliable patch neighbourhood system. Moreover, 
we will attempt to distinguish and extract the common semantic objects in a multi-class image set by combining object shape cues and semantic label cues. We believe that  multi-class objects co-segmentation will become an interesting and meaningful research direction in future image co-segmentation.

\bibliographystyle{IEEEtran}
\bibliography{paper}

\balance

\end{document}